\documentclass[sn-mathphys-num]{sn-jnl}

\usepackage{graphicx}%
\usepackage{multirow}%
\usepackage{amsmath,amssymb,amsfonts}%
\usepackage{amsthm}%
\usepackage{mathrsfs}%
\usepackage[title]{appendix}%
\usepackage{xcolor}%
\usepackage{textcomp}%
\usepackage{manyfoot}%
\usepackage{booktabs}%
\usepackage{algorithm}%
\usepackage{algorithmicx}%
\usepackage{algpseudocode}%
\usepackage{listings}%
\usepackage{bbm}
\usepackage{wrapfig}

\graphicspath{{graphics/}}

\begin{document}

\title[Article Title]{Enhancing Fractional Gradient Descent with Learned Optimizers}

\author*{\fnm{Jan} \sur{Sobotka}*}\email{sobotj11@fit.cvut.cz}
\author{\fnm{Petr} \sur{Šimánek}}\email{petr.simanek@fit.cvut.cz}
\author{\fnm{Pavel} \sur{Kordík}}\email{pavel.kordikk@fit.cvut.cz}

\affil{\orgdiv{Faculty of Information Technology}, \orgname{Czech Technical University in Prague}, \orgaddress{\street{Thákurova 9}, \city{Prague}, \postcode{16000}, \country{Czech Republic}}}

\abstract{
    Fractional Gradient Descent (FGD) offers a novel and promising way to accelerate optimization by incorporating fractional calculus into machine learning. Although FGD has shown encouraging initial results across various optimization tasks, it faces significant challenges with convergence behavior and hyperparameter selection. Moreover, the impact of its hyperparameters is not fully understood, and scheduling them is particularly difficult in non-convex settings such as neural network training. To address these issues, we propose a novel approach called Learning to Optimize Caputo Fractional Gradient Descent (L2O-CFGD), which meta-learns how to dynamically tune the hyperparameters of Caputo FGD (CFGD). Our method's meta-learned schedule outperforms CFGD with static hyperparameters found through an extensive search and, in some tasks, achieves performance comparable to a fully black-box meta-learned optimizer. L2O-CFGD can thus serve as a powerful tool for researchers to identify high-performing hyperparameters and gain insights on how to leverage the history-dependence of the fractional differential in optimization.
}


\maketitle

\section{Introduction}\label{sec:introduction}
    Recent advancements in machine learning have led to the exploration of novel techniques to improve optimization algorithms. One such approach involves the integration of fractional calculus principles into traditional gradient descent methods. This class of techniques, generally known as \textit{fractional gradient descent} (FGD), has shown promising empirical as well as theoretical results in accelerating the optimization process \cite{WANG201719,liu2023novel,WEI20202514,shin2021caputo,SHIN2023185}. However, despite its potential benefits, FGD faces significant challenges that hinder its practical utility. Namely, its convergence point is not the minimum point in the traditional sense due to the nonlocal property of the fractional-order differential \cite{WANG201719,frac2013}. Moreover, the speed of this method is highly dependent on the particular choice of the fractional order and other hyperparameters whose effects are not fully understood. For a recent and general review of various intersections of fractional calculus and machine learning, we refer the reader to \cite{e25010035}.

    On the opposite side of the spectrum of optimization methods are fully meta-learned approaches. Similarly to the overall trend in machine learning towards the replacement of manual feature engineering with learned features, there is a growing number of proposals on how to apply such a perspective to optimization. More specifically, the meta-learning subfield called \textit{learning to optimize} (L2O) has the ambitious goal of learning the optimization strategy itself, more or less replacing traditional hand-engineered optimizers such as (stochastic) gradient descent and Adam \cite{kingma2017adam}. Although the initial results of these data-driven methods seem promising, problems with their stability and generalization still remain, limiting their practical use \cite{10.5555/3157382.3157543,lv2017learning,metz2019understanding,harrison2022closer,Simanek_2022}.

    To circumvent the issues of FGD and enable its further progress, we propose an innovative method called \textit{Learning to Optimize Caputo Fractional Gradient Descent} (L2O-CFGD) that combines the adaptive nature of L2O to select hyperparameters for CFGD and tune them throughout the optimization run. On a range of both convex and non-convex tasks, L2O-CFGD accelerates the optimization process over classical CFGD and displays surprising yet interpretable behavior. These findings demonstrate that L2O-CFGD can serve as a powerful tool for researchers studying and developing fractional gradient descent. To the best of our knowledge, this is also the first attempt at finding a synergy between learning-based and fractional-calculus-based optimization methods.

\section{Background}\label{sec:background}
    We consider the general unconstrained optimization problem:
    \begin{equation}\label{eq:optimization-problem}
        \min_{\mathbf{x} \in \mathbb{R}^d} f(\mathbf{x}),
    \end{equation}
    where $f(\mathbf{x})$ is the objective function from $\mathbb{R}^d$ to $\mathbb{R}$. Furthermore, we focus on optimization methods that, starting at some initial point $\mathbf{x}^{(0)}$, make an update to $\mathbf{x}^{(t)}$ in iteration $t$ as
    \begin{equation}\label{eq:finding-solution}
        \mathbf{x}^{(t + 1)} = \mathbf{x}^{(t)} - \eta^{(t)} \cdot \mathbf{g}^{(t)},
    \end{equation}
    where $\eta^{(t)} \in \mathbb{R}$ is the learning rate and $\mathbf{g}^{(t)}$ is the update.

    \subsection{Caputo Fractional Derivative}
        There is no unified way to generalize the conventional derivative of integer order to the domain of real numbers. In this work, we consider one possible generalization known as the Caputo derivative \cite{10.1111/j.1365-246X.1967.tb02303.x}. Specifically, assuming that the function $f$ has at least $n+1$ continuous bounded derivatives in $[c, \infty)$, the Caputo fractional derivative of $f$ of order $\alpha$ is defined as \cite{Podlubny1999}
        \begin{equation}
            _c^C \mathcal{D}_x^\alpha f(x) =
                \frac{1}{\Gamma(n - \alpha)}
                \int_c^x \frac{f^{(n)}(\tau)}{(x - \tau)^{\alpha - n + 1}} \text{d}\tau,
        \end{equation}
        where $c \in \mathbb{R}$ is the integral terminal, $n \in \mathbb{N}_0$, $\alpha \in (n-1, n]$, and $\Gamma(\cdot)$ is the Gamma function.

        Intuitively, the Caputo fractional derivative induces an implicit regularization effect which, when applied in \eqref{eq:finding-solution} as the update, leads to the steepest descent direction of a certain smoothing of the original objective function $f$ \cite{shin2021caputo,SHIN2023185}. 

    \subsection{Caputo Fractional Gradient Descent}
    \label{sec:background-cfgd}
        Naturally, having defined a fractional derivative, we can now also generalize the standard gradient $\nabla f(\mathbf{x})$. Following \cite{shin2021caputo}, we define the Caputo fractional gradient of $f$ at the solution point $\mathbf{x}$ as follows.

        Let $f(\cdot)$ be a sufficiently smooth function from $\mathbb{R}^d$ to $\mathbb{R}$, $\mathbf{x} = (x_1,\dots,x_d) \in \mathbb{R}^d$, $j = 1,\dots,d$, and the function $f_{j,\mathbf{x}}: \mathbb{R} \rightarrow \mathbb{R}$ defined as
        \begin{align}
            f_{j,\mathbf{x}}(y) = f(\mathbf{x} + (y - x_j)\mathbf{e}_j),
        \end{align}
        where $\mathbf{e}_j$ denotes the $j$-th standard basis vector. Then, for vectors $\mathbf{c} = (c_1,\dots,c_d) \in \mathbb{R}^d$ and $\boldsymbol{\alpha} = (\alpha_1,\dots,\alpha_d) \in (0, 1]^d$, the Caputo fractional gradient of $f$ at $\mathbf{x}$ is defined as
        \begin{equation}
            _\mathbf{c}^C \nabla_\mathbf{x}^{\boldsymbol{\alpha}} f(\mathbf{x}) =
                \bigg(\,
                    _{c_1}^C \mathcal{D}_{x_1}^{\alpha_1} f_{1,\mathbf{x}}(x_1),\; \dots,\;
                    _{c_d}^C \mathcal{D}_{x_d}^{\alpha_d} f_{d,\mathbf{x}}(x_d)
                \bigg)^\text{T}
                \in \mathbb{R}^d.
        \end{equation}

        To introduce the full Caputo fractional gradient descent method (CFGD) \cite{shin2021caputo,SHIN2023185}, a properly scaled version of the Caputo fractional gradient is given by
        \begin{equation}\label{eq:caputo-fractional-gradient}
            _\mathbf{c} \mathbf{D}_{\boldsymbol{\beta}}^{\boldsymbol{\alpha}} f(\mathbf{x}) =
                \text{diag} \big(
                    \,_{c_j}^C \mathcal{D}_{x_j}^{\alpha_j} I(x_j)
                \big)^{-1}
                \bigg(\,
                    _\mathbf{c}^C \nabla_\mathbf{x}^{\boldsymbol{\alpha}} f(\mathbf{x}) +
                    \boldsymbol{\beta}\cdot
                    \text{diag}(|x_j - c_j|)\, _\mathbf{c}^C \nabla_\mathbf{x}^{1 + \boldsymbol{\alpha}} f(\mathbf{x})
                \bigg)
        \end{equation}
        where $\boldsymbol{\beta} = (\beta_1, \dots,\beta_d) \in \mathbb{R}^d$ are the smoothing parameters, $I$ is the identity map $I(x) = x$, and for a vector $\mathbf{v} = (v_1,\dots,v_d) \in \mathbb{R}^d$, $\text{diag}(v_j)$ or $\text{diag}(\mathbf{v})$ denotes the diagonal matrix of size $d \times d$ whose $(j,j)$ component is $v_j$. 

        Furthermore, when $f$ is a quadratic objective function in the form
        \begin{equation}\label{eq:quadratic-objective}
            f(\mathbf{x}) =
                \frac{1}{2} \mathbf{x}^\text{T} \mathbf{A} \mathbf{x} + \mathbf{b}^\text{T} \mathbf{x},
        \end{equation}
        where $\mathbf{A} \in \mathbb{R}^{d \times d}$ is a symmetric positive definite matrix and $\mathbf{b} \in \mathbb{R}^d$, the Caputo fractional gradient from \eqref{eq:caputo-fractional-gradient} can be expressed in a closed form
        \begin{equation}
            _\mathbf{c} \mathbf{D}_{\boldsymbol{\beta}}^{\boldsymbol{\alpha}} f(\mathbf{x}) =
                    \mathbf{A} \mathbf{x} + \mathbf{b} +
                    \text{diag}(\boldsymbol{\gamma}_{\boldsymbol{\alpha}, \boldsymbol{\beta}})
                    \text{diag}(\mathbf{r})(\mathbf{x} - \mathbf{c}).
        \end{equation}
        In the expression above, $\boldsymbol{\gamma}_{\boldsymbol{\alpha},\boldsymbol{\beta}} \in \mathbb{R}^d,\,(\boldsymbol{\gamma}_{\boldsymbol{\alpha},\boldsymbol{\beta}})_j = \beta_j - \frac{1 - \alpha_j}{2 - \alpha_j}$, and $\mathbf{r} = (\mathbf{A}_{1,1},\dots, \mathbf{A}_{d,d})^\text{T} \in \mathbb{R}^d$ is the diagonal of $\mathbf{A}$.

        Additionally, as shown in \cite{shin2021caputo}, for $\boldsymbol{\alpha} \in (0, 1)^d$ and $\boldsymbol{\beta}, \mathbf{c} \in \mathbb{R}^d$, the $j$-th component of $_\mathbf{c} \mathbf{D}_{\boldsymbol{\beta}}^{\boldsymbol{\alpha}} f(\mathbf{x})$ can be expressed as
        \begin{align}\label{eq:cfgd-integral}
            (_\mathbf{c} \mathbf{D}_{\boldsymbol{\beta}}^{\boldsymbol{\alpha}} f(\mathbf{x}))_j =\ 
                & C_j \int_{-1}^1
                    f_{j,\mathbf{x}}'(\Delta_j (1 + u) + c_j)(1 - u)^{-\alpha_j} \text{d}u \\
                & + C_j \beta_j |x_j - c_j| \int_{-1}^1
                    f_{j,\mathbf{x}}''(\Delta_j (1 + u) + c_j)(1 - u)^{-\alpha_j} \text{d}u, \notag
        \end{align}
        where $\Delta_j = \frac{|x_j - c_j|}{2}$ and $C_j = (1 - \alpha_j){2^{-(1-\alpha_j)}}$.

        With this in place, the Caputo fractional gradient descent updates the $t$-th iterated solution $\mathbf{x}^{(t)}$ of some optimization problem with the objective function $f$ as follows:
        \begin{equation}
            \mathbf{x}^{(t + 1)} = \mathbf{x}^{(t)} - \eta^{(t)}\cdot\,
            _\mathbf{c} \mathbf{D}_{\boldsymbol{\beta}}^{\boldsymbol{\alpha}} f(\mathbf{x}^{(t)}).
        \end{equation}

        When the hyperparameters $\boldsymbol{\alpha}$, $\boldsymbol{\beta}$, and $\mathbf{c}$ vary over time, we write them with a superscript $^{(t)}$.

        We note that the CFGD presented above, taken from \cite{shin2021caputo} and used in our experiments, differs in its formulation from that of \cite{SHIN2023185}. We were unable to replicate the results of the formulation found in \cite{SHIN2023185}.

    \subsubsection{Variants of Caputo Fractional Gradient Descent}
        Authors of \cite{shin2021caputo} proposed three variants of the CFGD algorithm described in \ref{sec:background-cfgd}. We limit ourselves to two of them: the non-adaptive CFGD (NA-CFGD) and adaptive-terminal CFGD (AT-CFGD).

        NA-CFGD sets all three hyperparameters to some constants that remain the same throughout all time steps $t$. Moreover, $\alpha_j$ and $\beta_j$ are the same across all components $j$. That is, $\boldsymbol{\alpha}^{(t)} = (\alpha, \dots, \alpha)$, $\boldsymbol{\beta}^{(t)} = (\beta, \dots, \beta)$, $\mathbf{c}^{(t)} = \mathbf{c}$, for some $\alpha \in (0, 1)$, $\beta \in \mathbb{R}$ and $\mathbf{c} \in \mathbb{R}^d$.

        Similarly, AT-CFGD sets $\boldsymbol{\alpha}^{(t)} = (\alpha, \dots, \alpha)$, $\boldsymbol{\beta}^{(t)} = (\beta, \dots, \beta)$ for some $\alpha \in (0, 1)$ and $\beta \in \mathbb{R}$, but uses an adaptive parameter $\mathbf{c}^{(t)} = \mathbf{x}^{(t-L)}$ for some positive integer $L$. Thus, AT-CFGD starts with predefined $L$ initial points $\{ \mathbf{x}^{(-k)} \}_{k=0}^L \subset \mathbb{R}^d$, maintaining a moving history of $\mathbf{x}$ throughout the optimization run.

    \subsection{Learning to Optimize}
    \label{sec:methods-learning-to-optimize}
        For the L2O method, we consider the architectural design introduced by \cite{10.5555/3157382.3157543}. The core idea is to use a recurrent neural network $M$, parameterized by $\phi$, that acts as the optimizer. Specifically, at each time step $t$, this network takes its hidden state $\mathbf{h}^{(t)}$ together with the gradient $\nabla_{\mathbf{x}} f(\mathbf{x}^{(t)})$ and produces an update $\mathbf{g}^{(t)}$ and a new hidden state $\mathbf{h}^{(t + 1)}$,
        \begin{equation}\label{eq:l2o-update}
            [\mathbf{g}^{(t)}, \mathbf{h}^{(t + 1)}] = M \big( \nabla_{\mathbf{x}} f(\mathbf{x}^{(t)}), \mathbf{h}^{(t)}, \phi \big).
        \end{equation}

        By applying these updates $\mathbf{g}^{(t)}$, the sequence of $\mathbf{x}^{(t)}$ obtained from \eqref{eq:finding-solution} then aims to converge to some local minimum of $f$. Therefore, in this context, $M$ is called the optimizer, or meta-learner, and $f$ is called the optimizee. We will use the term optimizee interchangeably with the objective function.

        During the meta-training phase, $\phi$ are learned using some variant of stochastic gradient descent, such as Adam, and updated every $u$-th inner optimization step where the hyperparameter $u$ is called the unroll. The loss of the optimizer is the weighted sum of the unrolled trajectory of the optimizee,
        \begin{equation}\label{eq:optimizee-train}
            \mathcal{L}(\phi) = \sum_{\tau=1}^{u} w^{(\tau)} f(\mathbf{x}^{(\tau + ju - 1)})
        \end{equation}
        where $w^{(\tau)}$ are weights that are typically set to 1. Furthermore, $j$ denotes the number of previously unrolled trajectories; therefore, $(j+1)$-th unrolled trajectory corresponds to training steps $t = ju, ju + 1, \dots, (j+1)u - 1$.

        In addition to updates of $\phi$ during a single optimization run (an inner loop), there is also an outer loop where the entire optimizee training is restarted from some initial $\mathbf{x}^{(t=0)}$ while the parameters $\phi$ continue to learn. In particular, the outer loop takes place only during the meta-training phase, where the main goal is to learn a good set of parameters $\phi$. The evaluation of the learned optimizer is then performed during meta-testing where $\phi$ is fixed and only $\mathbf{x}$ is updated.

        In practice, when there are several thousand or more parameters in $\mathbf{x}$, it is almost impossible to apply a general recurrent neural network. The authors of \cite{10.5555/3157382.3157543} avoid this issue by implementing the update rule coordinate-wise using a two-layer LSTM network with shared parameters. This means that the optimizer $M$ is a small network with multiple instances that share parameters $\phi$ but operate on distinct coordinates $j$ of the solution vector $\mathbf{x}$. For further algorithmic details and preprocessing, we refer the reader to \cite{10.5555/3157382.3157543}.

\section{Our Method}
\label{sec:methods}

    \subsection{Learning to Optimize Caputo Fractional Gradient Descent}
        Here, we propose an algorithm that combines L2O and CFGD called Learning to Optimize Caputo Fractional Gradient Descent (L2O-CFGD).

        The idea is to apply updates $\mathbf{g}^{(t)}$ from CFGD whose hyperparameters $\boldsymbol{\alpha}^{(t)}$, $\boldsymbol{\beta}^{(t)}$, and $\mathbf{c}^{(t)}$ are dynamically adjusted by the learned optimizer:
        \begin{align}\label{eq:l2o-cfgd-update}
            [\boldsymbol{\alpha}^{(t)}, \boldsymbol{\beta}^{(t)}, \mathbf{c}^{(t)}, \mathbf{h}^{(t + 1)}] &= M \big( \nabla_{\mathbf{x}} f(\mathbf{x}^{(t)}), \mathbf{h}^{(t)}, \phi \big) \\
            \mathbf{g}^{(t)} &=\, _\mathbf{c^{(t)}} \mathbf{D}_{\boldsymbol{\beta}^{(t)}}^{\boldsymbol{\alpha}^{(t)}} f(\mathbf{x}^{(t)}).
        \end{align}

        In this way, it is not necessary to manually tune the fractional order $\boldsymbol{\alpha}$ or the CFGD hyperparameters $\boldsymbol{\beta}$ and $\mathbf{c}$. Notice that this update rule allows for the same meta-training of the learned optimizer as described in \ref{sec:methods-learning-to-optimize} and is described in full detail in Algorithm \ref{algo:cfgd}.

    \begin{algorithm}[h]
        \caption{L2O-CFGD meta-training on a general objective function}\label{algo:cfgd}
        \begin{algorithmic}[1]
            \Require
                \Statex Meta-optimizer $\hat{M}$ \Comment{Any variant of SGD}
                \Statex Optimizee $f$
                \Statex Optimizee learning rate $\eta$
                \Statex Number of meta-training optimization runs $N_{\text{meta}}$
                \Statex Unroll $u$
                \Statex Maximum time step $T$
                \Statex Number of points for the Gauss-Jacobi quadrature $s$
                \Statex Number of Hutchinson steps for evaluating $_\mathbf{c} \mathbf{Q}_{\boldsymbol{\beta}}^{\boldsymbol{\alpha}} f(\mathbf{x})$
    
            \Ensure Learned parameters $\phi$
            \State Initialize $\phi$
            \For{$k \gets 1$ to $N_{\text{meta}}$}
                \State Set unroll loss $\mathcal{L}_u \gets 0$
                \State Initialize $\mathbf{x}^{(t=0)}, \mathbf{h}^{(t=0)}$
                \For{$t \gets 0$ to $T-1$}
                    \State $[\boldsymbol{\alpha}^{(t)}, \boldsymbol{\beta}^{(t)}, \mathbf{c}^{(t)}, \mathbf{h}^{(t + 1)}] \gets M \big( \nabla_{\mathbf{x}} f(\mathbf{x}^{(t)}), \mathbf{h}^{(t)}, \phi \big)$
                    \State Compute $\{ (v_{j,l}, w_{j,l}) \}_{l=1}^s$ for each component $j$ in $\boldsymbol{\alpha}^{(t)}$
                    \State $\mathbf{g}^{(t)} \gets \, _\mathbf{c^{(t)}} \mathbf{Q}_{\boldsymbol{\beta}^{(t)}}^{\boldsymbol{\alpha}^{(t)}} f(\mathbf{x}^{(t)})$ \Comment{Using $\{ (v_{j,l}, w_{j,l}) \}_{l=1}^s$ from previous step}
                    \State $\mathbf{x}^{(t+1)} \gets \mathbf{x}^{(t)} - \eta \cdot \mathbf{g}^{(t)}$
    
                    \State $\mathcal{L}_u \gets \mathcal{L}_u + f(\mathbf{x}^{(t+1)})$
                    \If{$(t+1) \bmod u = 0$}
                        \State Backpropagate $\mathcal{L}_u$ to $\phi$ and update $\phi$ using $\hat{M}$
                        \State $\mathcal{L}_u \gets 0$
                    \EndIf
                \EndFor
            \EndFor
        \end{algorithmic}
        \end{algorithm}

    \subsection{Approximating the Caputo Fractional Gradients}
        Authors of \cite{shin2021caputo} already observed that \eqref{eq:cfgd-integral} involves an integral that can be accurately evaluated by the Gauss-Jacobi quadrature. That is, unless there is a closed form for $_\mathbf{c} \mathbf{D}_{\boldsymbol{\beta}}^{\boldsymbol{\alpha}} f(\mathbf{x})$, we use the Gauss-Jacobi quadrature rule of $s$ points $\{ (v_{j,l}, w_{j,l}) \}_{l=1}^s$, corresponding to the order $\alpha_j$, to approximate $(_\mathbf{c} \mathbf{D}_{\boldsymbol{\beta}}^{\boldsymbol{\alpha}} f(\mathbf{x}))_j$ by
        \begin{align}\label{eq:cfgd-quad}
            (_\mathbf{c} \mathbf{Q}_{\boldsymbol{\beta}}^{\boldsymbol{\alpha}} f(\mathbf{x}))_j =
                & \ C_j \sum_{l=1}^s w_{j,l}
                f_{j,\mathbf{x}}'(\Delta_j (1 + v_{j,l}) + c_j) \\
                & + C_j \beta_j |x_j - c_j|
                \sum_{l=1}^s w_{j,l}
                f_{j,\mathbf{x}}''(\Delta_j (1 + v_{j,l}) + c_j). \notag
        \end{align}
        As empirically observed in \cite{shin2021caputo,SHIN2023185} and in our experiments, setting $s=1$ is sufficient for many practical problems and leads to only a very minor change in performance.
        
        Since \eqref{eq:cfgd-quad} requires the diagonal components of the Hessian and our considered optimization problems involve thousands of parameters, we opt to approximate it using the Hutchinson's method procedure described in \cite{Yao_Gholami_Shen_Mustafa_Keutzer_Mahoney_2021}.

        The idea is to use an oracle to compute the multiplication between the Hessian matrix $\mathbf{H}$ and a random vector $\mathbf{z}$ without explicitly forming the full Hessian, requiring only the gradient $\mathbf{g}$:
        \begin{equation}\label{eq:hessian-diagonal-oracle}
            \frac{\partial \mathbf{g}^\text{T} \mathbf{z}}{\partial \mathbf{x}} =
            \frac{\partial \mathbf{g}^\text{T}}{\partial \mathbf{x}} \mathbf{z} +
            \mathbf{g}^\text{T} \frac{\partial \mathbf{z}}{\partial \mathbf{x}} =
            \frac{\partial \mathbf{g}^\text{T}}{\partial \mathbf{x}} \mathbf{z} =
            \mathbf{H} \mathbf{z}.
        \end{equation}
        One can notice that this Hessian-free oracle requires only backpropagating the $\mathbf{g}^\text{T} \mathbf{z}$ term, which is efficiently implemented in many deep learning libraries. In turn, we can compute the Hessian diagonal using the Hutchinson's method as
        \begin{equation}
            \text{diag}(\mathbf{H}) = \mathbb{E}[\mathbf{z} \odot (\mathbf{H} \mathbf{z})],
        \end{equation}
        where $\mathbf{z}$ is a random vector with Rademacher distribution, and $\mathbf{H} \mathbf{z}$ is calculated using the Hessian-free oracle from \eqref{eq:hessian-diagonal-oracle}. The proof of this equality can be found in \cite{BEKAS20071214}.

\section{Results}\label{sec:results}
    The goal of our experiments is two-fold. First, we want to find out if L2O-CFGD can exceed the performance of CFGD with hyperparameters set from a hyperparameter search. Second, our aim is to uncover useful insights from the time-varying hyperparameter selection performed by L2O-CFGD. Together, the experiments will assess whether L2O-CFGD can be a valuable tool for studying and advancing fractional gradient descent methods.
    
    To achieve these objectives, we present four optimization problems and compare the performance of Gradient Descent (GD), NA-CFGD, AT-CFGD, fully black-box L2O, and L2O-CFGD. The selection of problems is based on previous work \cite{shin2021caputo,SHIN2023185}, limited computational resources, and the aim of evaluating L2O-CFGD in both convex and non-convex settings.
    
    In all experiments, we include an additional linear encoding of the time step $t$ in the input of the L2O and L2O-CFGD optimizer networks (scalar value in the range $[0,1]$). We observed that this auxiliary input leads to better results for both methods.
    
    Both meta-learning methods are meta-trained using the Adam optimizer with a learning rate of $0.001$, and the optimizee learning rate is set to $0.1$. Additionally, for L2O-CFGD, we used 3 Hutchinson steps per iteration.

    \subsection{Quadratic Objective Function}\label{sec:results-quadratic}
        Here we consider the least squares problem formulated through the objective function
        \begin{equation}
            f(\mathbf{x}) = \frac{1}{2} || \mathbf{W}^\text{T} \mathbf{x} - \mathbf{y}||^2,
        \end{equation}
        where $\mathbf{W} \in \mathbb{R}^{d \times m}$, $\mathbf{y} \in \mathbb{R}^{m}$, and $||\cdot||$ is the Euclidean norm.
    
        It can be checked that using the line search in the equivalent quadratic formulation \eqref{eq:quadratic-objective} of the problem
        \begin{equation}
            \min_\eta \frac{1}{2} \mathbf{x}^{(t+1)} \mathbf{A} \mathbf{x}^{(t+1)} + \mathbf{b}^\text{T} \mathbf{x}^{(t+1)},
        \end{equation}
        where $\mathbf{A} = \mathbf{W} \mathbf{W}^\text{T}, \mathbf{b} = -\mathbf{W} \mathbf{y}^\text{T}$, the optimal learning rate for \eqref{eq:finding-solution} is given by
        \begin{equation}\label{eq:optimal-learning-rate}
            (\eta^{(t)})^{*} = \frac{
                \langle \mathbf{A} \mathbf{x}^{(t)} + \mathbf{b}, \mathbf{g}^{(t)} \rangle
            }{
                (\mathbf{g}^{(t)})^\text{T} \mathbf{A} \mathbf{g}^{(t)}
            }.
        \end{equation}
        This optimal learning rate is used for all the optimizers considered in this section.

        We meta-train both L2O-CFGD and L2O on 2000 runs of 800 iterations of the least squares optimization problem $d = m = 100$, with the unroll $u$ set to 20. Then, to check how sensitive the meta-learned optimization strategy is to the size of the problem, we evaluate the performance with three different settings of $d$ and $m$.
    
        \begin{figure}[b]
            \centering
                \includegraphics[width=\linewidth]{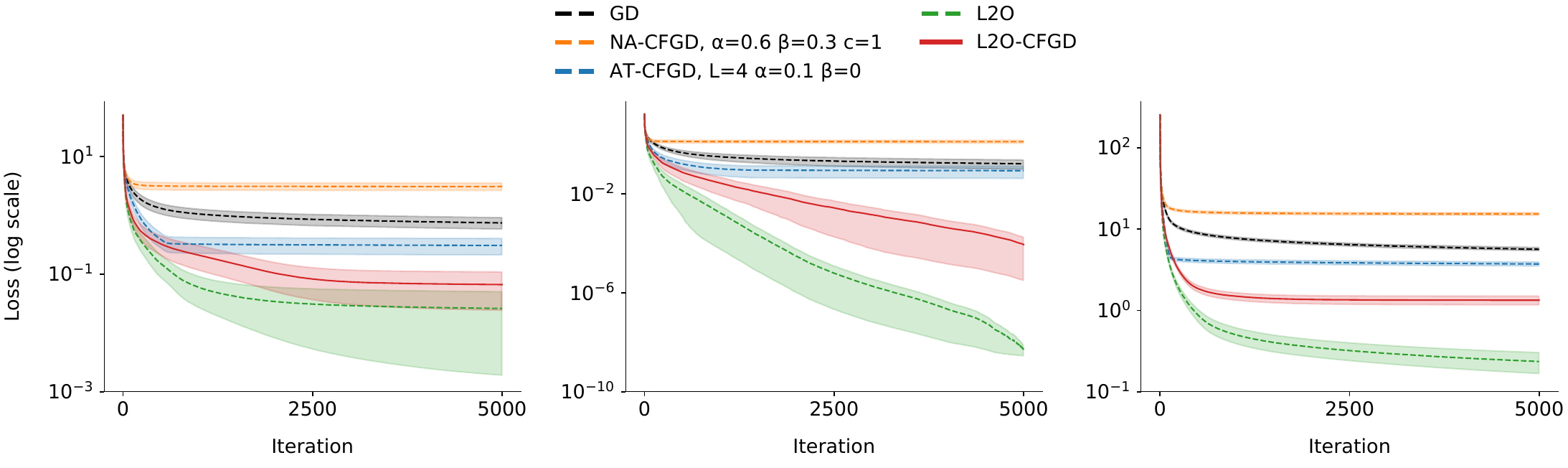}
                \caption{Comparison of GD, NA-CFGD, AT-CFGD, L2O and L2O-CFGD. Left: $d = m = 100$ as in meta-training of L2O-CFGD and L2O. Middle: $d = m = 30$. Right: $d = m = 500$. Performance averaged across 15 runs.}
                \label{fig:quadratic-loss}
        \end{figure}

        The hyperparameter search space of NA-CFGD was $\big\{(\alpha, \beta, c)\, | \, \alpha \in \{0.2,0.6,0.9\};\, \beta \in \{-5,-1,-0.3,0,0.3,1,5\};\, c \in \{-10,-1,-0.5,0,0.5,1,10\} \big\}$ from which we selected $\alpha=0.6$, $\beta=0.3$, $c=1$ as the best-performing combination. Similarly for AT-CFGD where we removed $c$ from the search space and added $\text{L} \in \{1,2,3,4\}$, leading to $\alpha=0.1$, $\beta=0$, $L=4$ as the best combination. As in the original work \cite{shin2021caputo}, the $\text{L}$ initial points were sampled from the standard normal distribution.

        The results are shown in Figure \ref{fig:quadratic-loss}. As we can see, in all three settings, L2O-CFGD outperforms the other two CFGD variants and is able to generalize across different sizes of the task. Moreover, we can inspect the strategy employed by L2O-CFGD to obtain some insights into its well-performing schedule for CFGD hyperparameters. To do so, we track the progression of the dynamically adjusted hyperparameters $\boldsymbol{\alpha}^{(t)}$, $\boldsymbol{\beta}^{(t)}$, and $\mathbf{c}^{(t)}$ during the meta-testing run on the $d = m = 100$ optimization problem.

\begin{figure}[h]
        \centering
            \includegraphics[width=\linewidth]{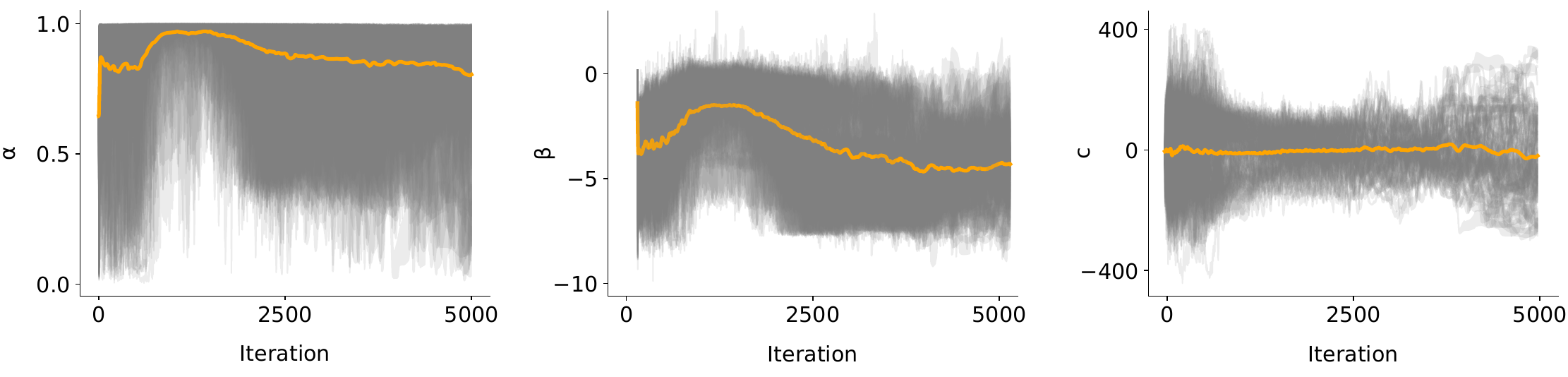}
            \caption{Progression of the dynamically adjusted hyperparameters $\boldsymbol{\alpha}^{(t)}$, $\boldsymbol{\beta}^{(t)}$, and $\mathbf{c}^{(t)}$ in the $d = m = 100$ optimization run of L2O-CFGD. The gray lines show the trajectories of $\alpha_j^{(t)}$, $\beta_j^{(t)}$, $c_j^{(t)}$ for different optimizee parameters $x_j$, and the orange line shows the mean across all coordinates $j$.}
            \label{fig:quadratic-strategy}
        \end{figure}

        As can be seen in Figure \ref{fig:quadratic-strategy}, the initial rapid drop in the loss is accompanied by a steep increase in fractional order $\alpha$ and a drop in $\beta$ in the first iterations. It is interesting that around the 500th iteration, when the loss reaches a more steady decrease and AT-CFGD with static hyperparameters starts to plateau, $\alpha$ and $\beta$ temporarily increase. This indicates that there might be some transient change in the optimization landscape that L2O-CFGD can deal with, as opposed to AT-CFGD which struggles to continue. From the progression of the hyperparameter $c$ which has a high variance between different optimizee parameters, we can deduce that, in this problem setup, a good performance of CFGD might require a highly coordinate and time-specific hyperparameter schedule.

    \subsection{Training Neural Networks}\label{sec:results-nn}
        To move from a convex to a more difficult and generally non-convex setting, we consider neural network training on three separate tasks.

        The first two tasks are defined through the following functions which the network needs to learn:
        \begin{align}
        h_1(z) &= \sin(2\pi z) e^{-z^2} \\
        h_2(z) &= \mathbbm{1}_{z > 0}(z) + 0.2 \cdot \sin(2\pi z),
        \end{align}
        and the last is an image classification task on the MNIST dataset with the cross-entropy loss function.

    \subsubsection{Functions $h_1$ and $h_2$}\label{sec:results-h12}
\begin{wrapfigure}{R}{0.35\textwidth}
            \includegraphics[width=0.35\textwidth]{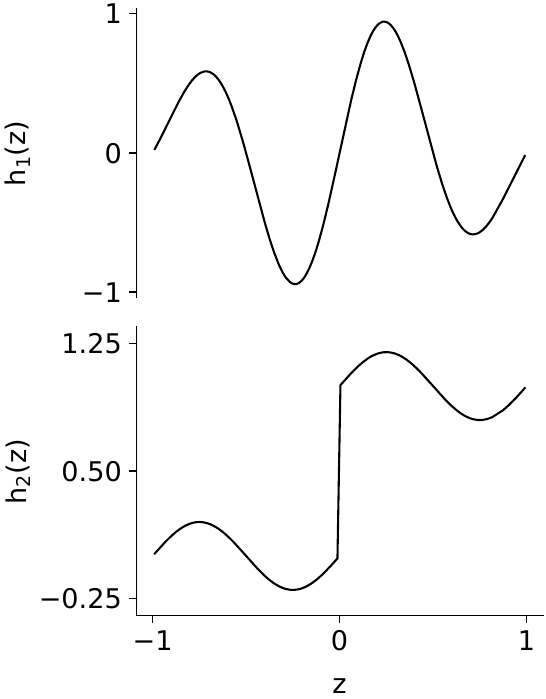} 
            \vspace{-10pt}
            \caption{Illustration of the test functions $h_1$ and $h_2$.}
            \label{fig:h-funcs}
            \vspace{-30pt}
        \end{wrapfigure}

        For the $h_1$ and $h_2$ functions, shown in Figure \ref{fig:h-funcs}, the optimizee is a univariate hyperbolic tangent neural network with 1 hidden layer of 50 neurons. The training dataset consists of 100 points $\{(z_i,h(z))\}_{i=1}^{100}$ where $z_i$ are sampled uniformly from the range [-1,1], and no mini-batching is performed. The objective (loss) function is the residual sum of squares.

        For all optimizers compared in this section, we perform a 1-step look-ahead learning rate search in each step. The set of learning rates tested is $\{t \cdot 10^{(-l)}\}$ where $t \in \{0.25, 0.5, 0.75, 1\}$ and $l \in \{1, \dots, 7\}$.

        The hyperparameter search space for NA-CFGD was
        \begin{align*}
        \big\{(\alpha, \beta, c)\, | \, &\alpha \in \{0.2,0.4,0.7,0.95\};\\ &\beta \in \{-50,-10,-1,0,1,10,50\};\\ &c \in \{-5,-1,-0.5,0,0.5,1,5\} \big\},
        \end{align*}
        
        from which combinations $\alpha=0.95$, $\beta=0$, $c=-5$ and $\alpha=0.95$, $\beta=0$, $c=-0.5$ performed best for $h_1$ and $h_2$, respectively. For AT-CFGD, we removed $c$ and included $\text{L} \in \{1,2,3,4\}$ in the search space, resulting in $\alpha=0.2$, $\beta=-5$, $\text{L}=1$ for $h_1$ and $\alpha=0.95$, $\beta=-1$, $\text{L}=1$ for $h_2$.
            
        L2O-CFGD and L2O are meta-trained on 1200 runs of length 600, with unroll $u=40$ for both $h_1$ and $h_2$. We observed no further improvement from a longer meta-training.

\begin{figure}[b]
        \centering
            \includegraphics[width=0.95\linewidth]{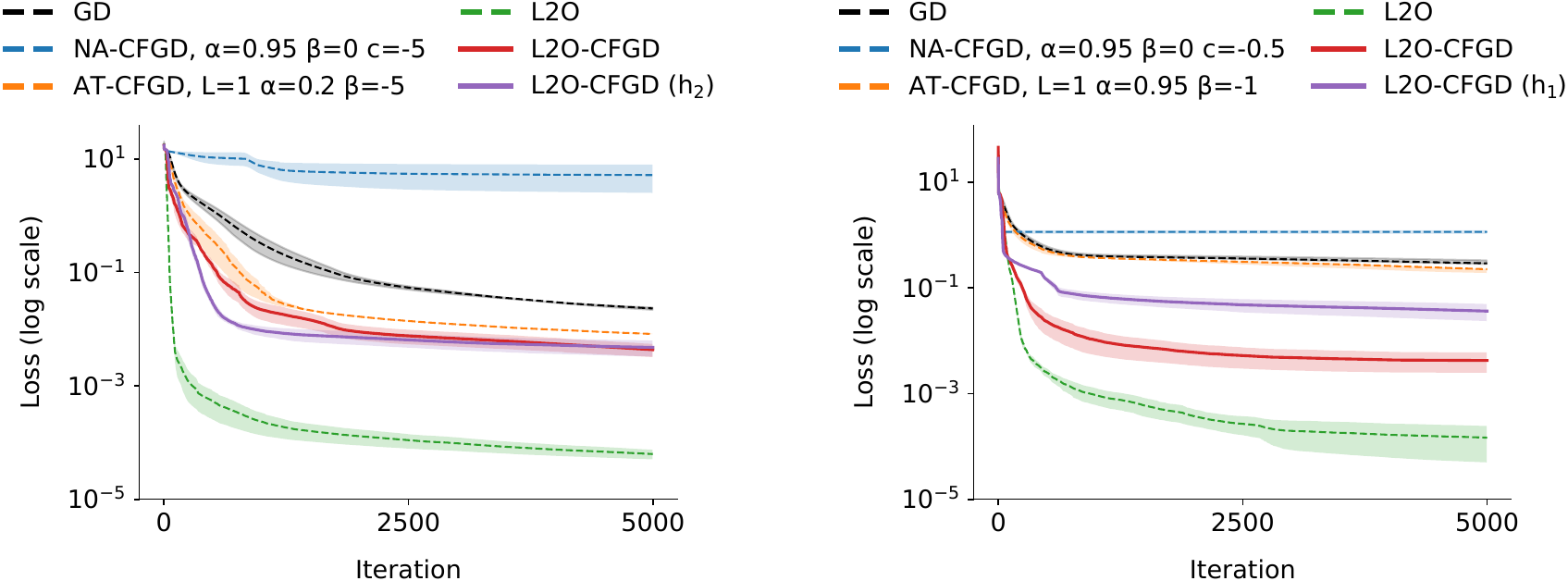}
            \caption{Comparison of GD, NA-CFGD, AT-CFGD, L2O and L2O-CFGD. The function name in brackets after L2O-CFGD denotes the function on which it was meta-trained. Left: The $h_1$ function. Right: The $h_2$ function. Performance averaged across 5 runs.}
            \label{fig:h-funcs-loss}
        \end{figure}

        In Figure \ref{fig:h-funcs-loss}, we plot both the performance of L2O-CFGD meta-trained on the given test function, as well as the performance of L2O-CFGD originally meta-trained on the other test function to validate the method's ability to generalize. One can see that, similarly to the least squares optimization task, L2O-CFGD outperforms both NA-CFGD and AT-CFGD, which points to its better CFGD hyperparameter schedule. L2O-CFGD also appears to generalize well; in fact, it is surprising that L2O-CFGD meta-trained on $h_2$ outperforms L2O-CFGD meta-trained on $h_1$ when evaluated on $h_1$ (Figure \ref{fig:h-funcs-loss}, left).

\begin{figure}[t]
        \centering
            \includegraphics[width=\linewidth]{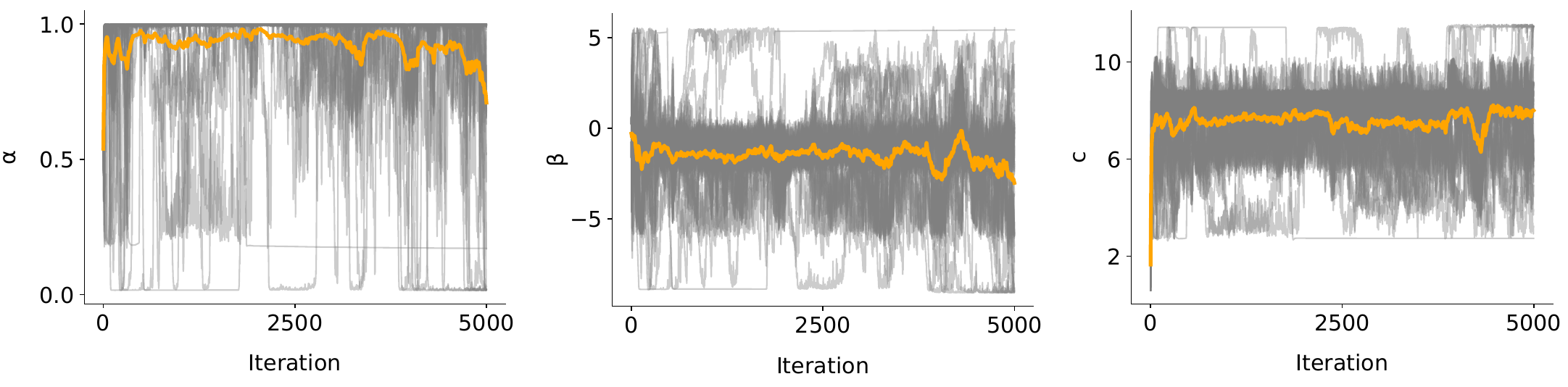}
            \caption{Progression of the dynamically adjusted hyperparameters $\boldsymbol{\alpha}^{(t)}$, $\boldsymbol{\beta}^{(t)}$, and $\mathbf{c}^{(t)}$ by L2O-CFGD for learning the $h_1$ function with a neural network. The gray lines show the trajectories of $\alpha_j^{(t)}$, $\beta_j^{(t)}$, $c_j^{(t)}$ for different optimizee parameters $x_j$, and the orange line shows the mean across all coordinates $j$.}
            \label{fig:h1-strategy}
        \end{figure}

\begin{figure}[t]
        \centering
            \includegraphics[width=\linewidth]{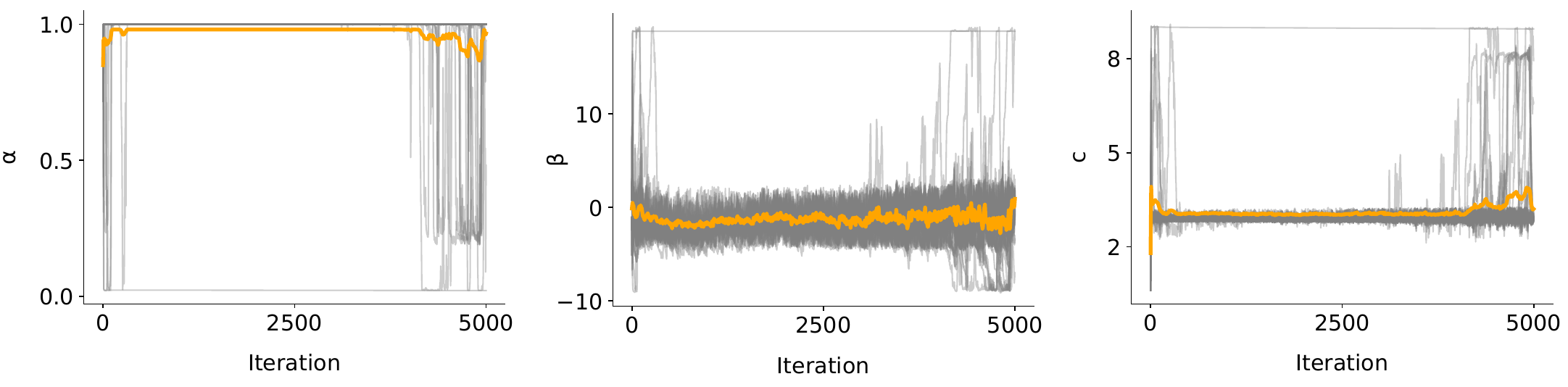}
            \caption{Progression of the dynamically adjusted hyperparameters $\boldsymbol{\alpha}^{(t)}$, $\boldsymbol{\beta}^{(t)}$, and $\mathbf{c}^{(t)}$ by L2O-CFGD for learning the $h_2$ function with a neural network. The gray lines show the trajectories of $\alpha_j^{(t)}$, $\beta_j^{(t)}$, $c_j^{(t)}$ for different optimizee parameters $x_j$, and the orange line shows the mean across all coordinates $j$.}
            \label{fig:h2-strategy}
        \end{figure}

        In Figures \ref{fig:h1-strategy} ($h_1$) and \ref{fig:h2-strategy} ($h_2$), we see that L2O-CFGD found $\alpha$ close to 1 and $\beta$ just below 0 to be a good hyperparameter combination. This agrees with the hyperparameter search of AT-CFGD and NA-CFGD in most cases, where similar values were chosen for $\alpha$ and $\beta$.

    \subsubsection{MNIST Classification Task}\label{sec:results-mnist}
        In this last problem setup, we meta-train L2O-CFGD and L2O on feed-forward neural network with 1 hidden layer of 20 neurons with the ReLU activation function. We put the softmax activation function at the output and train the network on the MNIST classification task with the cross-entropy loss function and batch size of 128. This forms the optimizee $f$ with parameters $\mathbf{x}$. We set the unroll $u$ to 40 iterations and perform meta-training for 1200 separate optimization runs with a maximum iteration number of 400. In each of the training runs, the optimizee parameters are randomly reinitialized from $\mathcal{U}(-\sqrt{k}, \sqrt{k})$ where $k = \frac{1}{\text{in features}}$ (PyTorch's default initialization).

        To test the generalization capability of the optimizers, we perform meta-testing on the optimizee architecture from meta-training, as well as on a larger network with two layers and three times as many neurons per layer.

        For SGD, we chose a learning rate of 0.3 from a hyperparameter search. Similarly for NA-CFGD and AT-CFGD where the search space was $\big\{(\alpha, \beta, c, \eta)\, | \, \alpha \in \{0.1,0.3,0.6,0.9\};\, \beta \in \{-20,-2,0,2,20\};\, c \in \{-1,0,1\};\, \eta \in \{0.003,0.02,0.04\} \big\}$ for NA-CFGD and $\big\{(\alpha, \beta,\text{L},\eta)\, | \, \alpha \in \{0.2,0.6,0.9\};\, \beta \in \{-10,-2,-1,0,1,2,10\};\, \text{L} \in \{1,2,3,4\};\,\eta \in \{0.1,0.5\} \big\}$ for AT-CFGD. The best-performing combinations were $\alpha=0.1$, $\beta=0$, $c=0$, $\eta=0.02$ (NA-CFGD) and $\alpha=0.2$, $\beta=-1$, $\text{L}=1$, $\eta=0.1$ (AT-CFGD).

        \begin{figure}[t]
        \centering
            \includegraphics[width=0.9\linewidth]{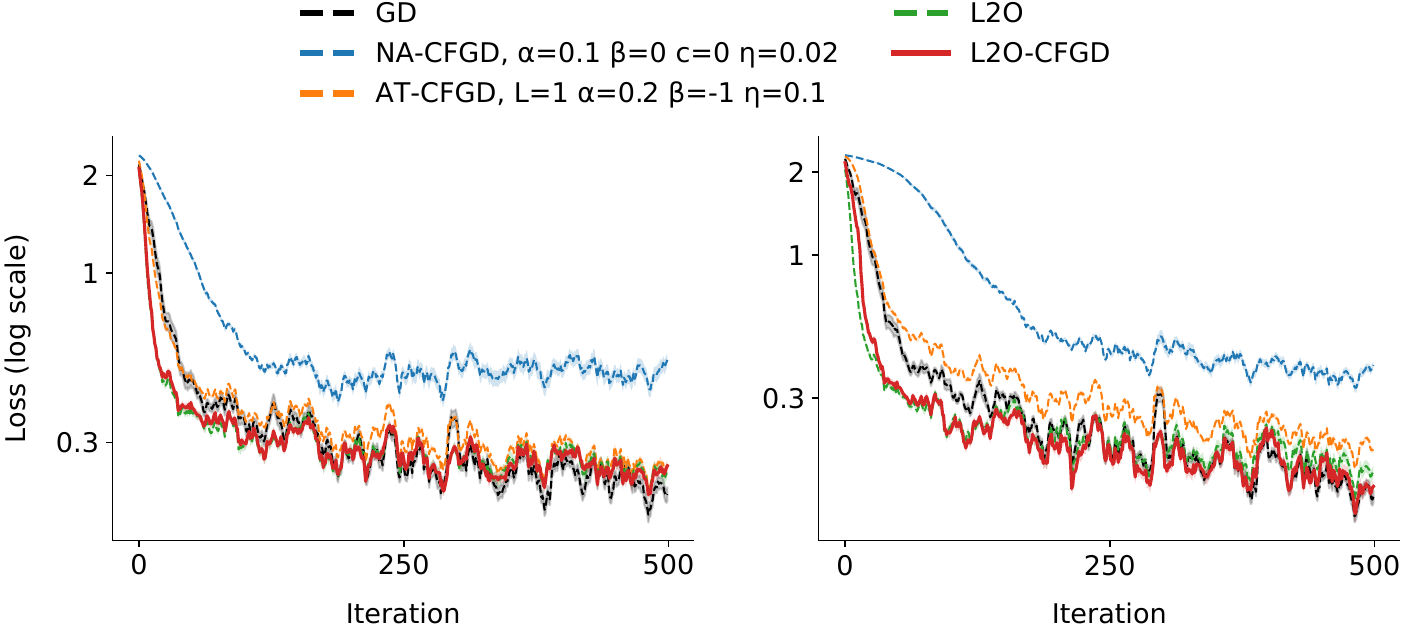}
            \caption{Comparison of SGD, NA-CFGD, AT-CFGD, L2O and L2O-CFGD. Left: Training the ReLU optimizee. Right: Training ReLU optimizee with two layers of 60 neurons each. Performance averaged across 10 runs.}
            \label{fig:nn-loss}
        \end{figure}

        From the results in Figure \ref{fig:nn-loss}, we see that neither NA-CFGD nor AT-CFGD can outperform SGD. On the other hand, L2O-CFGD and L2O are faster in the initial phase of training and achieve better results than both the adaptive-terminal and non-adaptive CFGD.  Furthermore, we can observe that L2O-CFGD almost perfectly matches the performance of L2O. 

        \begin{figure}[b]
            \centering
                \includegraphics[width=\linewidth]{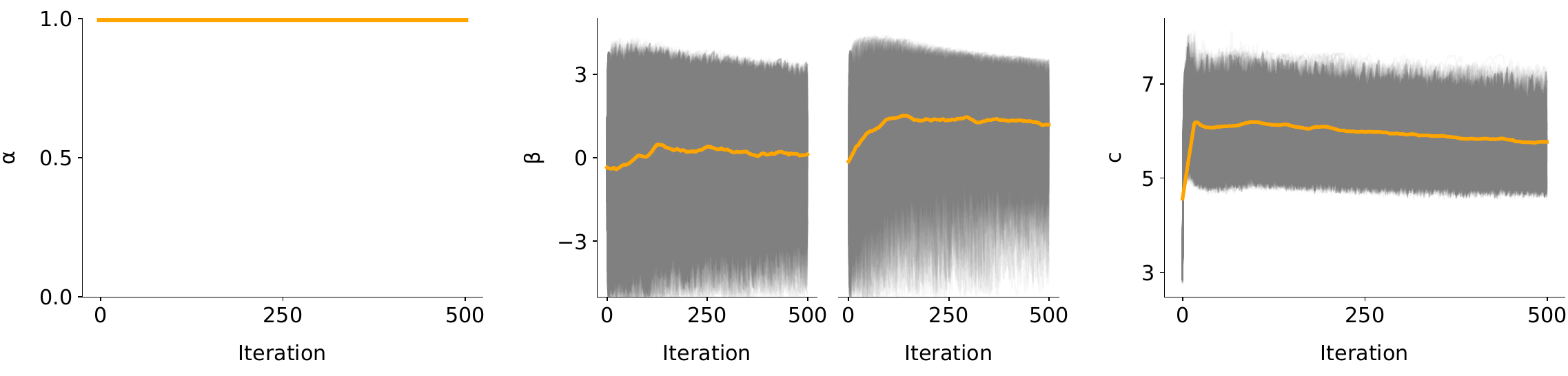}
                \caption{Progression of the dynamically adjusted hyperparameters $\boldsymbol{\alpha}^{(t)}$, $\boldsymbol{\beta}^{(t)}$, and $\mathbf{c}^{(t)}$ in the optimization run of L2O-CFGD on the ReLU optimizee. The gray lines show the trajectories of randomly sampled $\alpha_j^{(t)}$, $\beta_j^{(t)}$, $c_j^{(t)}$ for different optimizee parameters $x_j$, and the orange line shows the mean across all the tuned hyperparameters. Hyperparameters $\boldsymbol{\beta}^{(t)}$ are shown separately for coordinates $j$ in the mapping from inputs to the hidden representation (left) and from the hidden representation to the output (right).}
                \label{fig:nn-strategy}
        \end{figure}
        
        Regarding the optimization strategy behind L2O-CFGD, Figure \ref{fig:nn-strategy} illustrates that it has meta-learned to keep $\alpha$ fixed at 1 for all coordinates $j$, indicating its effectiveness for the given task. From the $\beta$ plots, we can conclude that separate $\beta$ schedules for the coordinates $j$ in the input-to-hidden and hidden-to-output parameter matrices are advantageous. This observation opens up an intriguing research direction to explore the relationship between neural network depth and the history-dependence required for the fractional differential used in optimizing the particular layer (governed by $\beta$).

\begin{figure}[b]
        \centering
            \includegraphics[width=\linewidth]{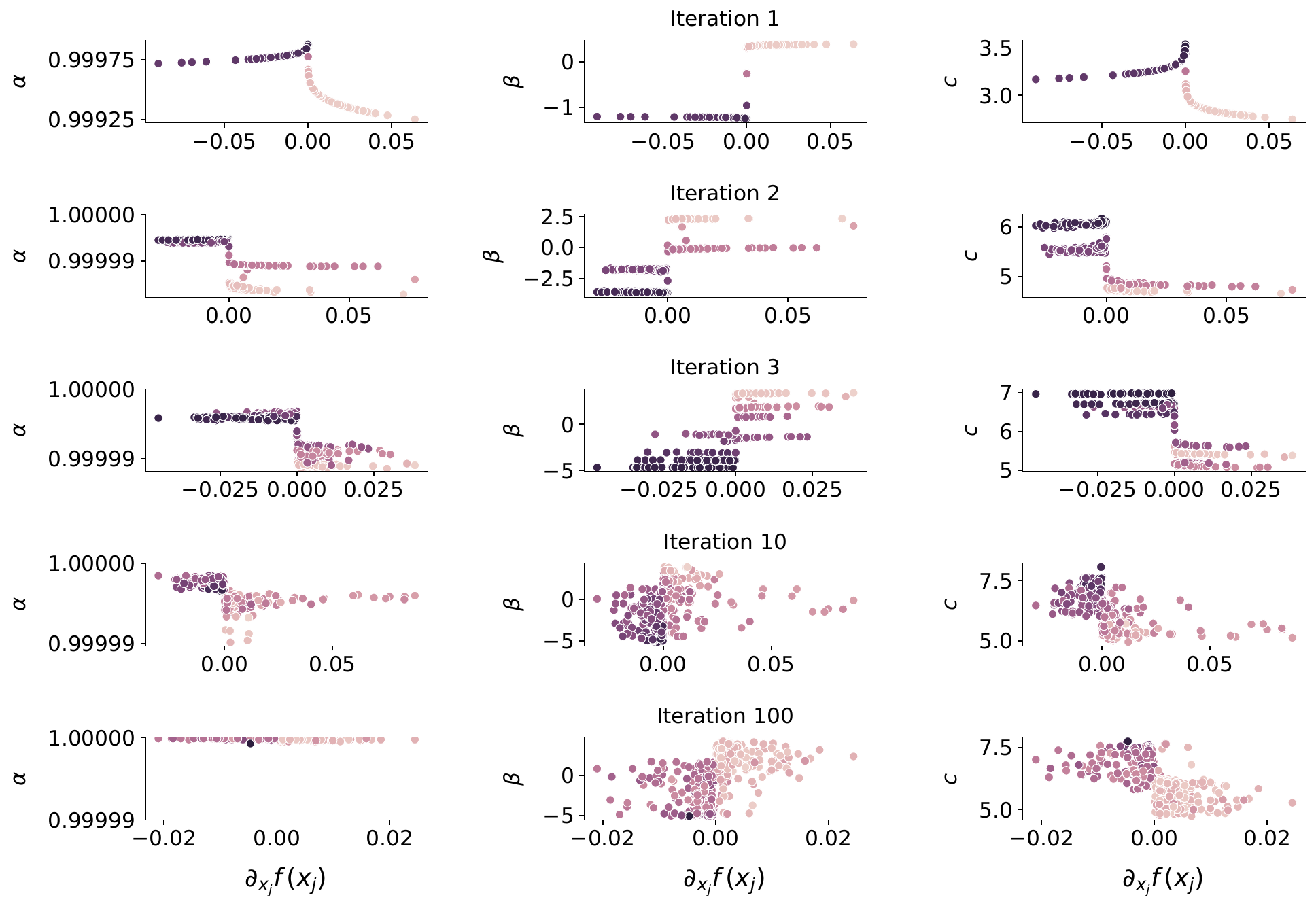}
            \caption{Progression of the dynamically adjusted hyperparameters $\alpha_j^{(t)}$, $\beta_j^{(t)}$, and $c_j^{(t)}$ over individual iterations in relation to the partial derivative w.r.t. the corresponding component $x_j$. The color indicates the mean value of the hidden state for the particular $x_j$ (darker color represents higher values). The data comes from the optimization run of L2O-CFGD on the ReLU optimizee. For visualization purposes, only a randomly sampled subset of the total components is shown.}
            \label{fig:iterations-strategy}
        \end{figure}

        In Figure \ref{fig:iterations-strategy}, we can further inspect how L2O-CFGD translates the partial derivatives of the objective function from its input into individual hyperparameters $\alpha_j^{(t)}$, $\beta_j^{(t)}$ and $c_j^{(t)}$ on its output. The color indicates the dependence on the hidden state. We plot only the first few iterations since after around the 100th iteration, the learned mapping shows very similar characteristics.
        
        As we can see, the meta-learned strategy starts with a separation of parameters into two groups: The parameters $x_j$ with positive partial derivatives get negligibly higher $\alpha_j^{(t)}$, significantly higher $\beta_j^{(t)}$ and lower $c_j^{(t)}$ than the parameters with negative partial derivatives. It also highly correlates with the mean value of the hidden state corresponding to the particular coordinate. Interestingly, we can observe that this separation fades away after the initial rapid drop in loss around the 100th iteration when the optimization enters the phase of a more gradual decrease in the loss.

        Overall, the results show that $\alpha=1$ is the right choice and that there exist general rules of thumb for the other hyperparameters. These findings further motivate the use of data-driven approaches, such as L2O, to uncover strategies and improvements to FGD methods.

\section{Conclusion}\label{sec:conclusion}
    In this study, we introduced Learning to Optimize Caputo Fractional Gradient Descent (L2O-CFGD), a novel approach that bridges fractional calculus and meta-learning to enhance the optimization process. Our method addresses the challenges of hyperparameter selection and convergence behavior in the Caputo fractional gradient descent (CFGD) by dynamically tuning hyperparameters throughout the optimization run. Experimental results demonstrate that L2O-CFGD not only outperforms traditional CFGD methods but also achieves performance comparable to fully black-box learned optimizers in certain neural network training tasks.

    Beyond its performance, L2O-CFGD offers valuable insights into the role of the hyperparameters and their scheduling in fractional gradient descent, highlighting the potential of leveraging the history-dependent properties of fractional differential in optimization. We believe that L2O-CFGD will serve as a powerful tool for researchers, facilitating the exploration of high-performing hyperparameters and advancing the understanding of fractional calculus in optimization.

\backmatter


\bmhead{Acknowledgements}
    This work was supported by the Student Summer Research Program 2023 of FIT CTU in Prague.


    

    



    \subsection*{Code Availability}
        Code is available at \url{https://github.com/Johnny1188/fractional-learning-to-optimize}.





\bibliography{sn-bibliography}

\end{document}